\documentclass[11pt,a4paper]{article}
\usepackage[hyperref]{acl2021}
\usepackage{times}
\usepackage{latexsym}

\usepackage{times}
\usepackage{latexsym}
\usepackage{graphicx}
\usepackage{url}
\usepackage{comment}
\usepackage{float}
\usepackage{amsmath}
\usepackage{algorithmic}
\usepackage[ruled,vlined]{algorithm2e}
\usepackage{multirow}
\usepackage{float}
\usepackage{microtype}

\aclfinalcopy 

\title{Event Detection as Question Answering with Entity Information}

\author{Emanuela Boros \\
  University of La Rochelle, L3i \\
  F-17000, La Rochelle, France  \\
  {\tt \small{emanuela.boros@univ-lr.fr}} \\\And
  Jose G. Moreno \\
  University of Toulouse, IRIT \\
  F-31000, Toulouse, France \\
  {\tt \small{jose.moreno@irit.fr}}  \\\And
  Antoine Doucet \\
  University of La Rochelle, L3i \\
  F-17000, La Rochelle, France  \\
  {\tt \small{antoine.doucet@univ-lr.fr}}
  \\}

\date{}

\begin{document}
\maketitle
\begin{abstract}

In this paper, we propose a recent and under-researched paradigm for the task of event detection (ED) by casting it as a question-answering (QA) problem with the possibility of multiple answers and the support of entities. The extraction of event triggers is, thus, transformed into the task of identifying answer spans from a context, while also focusing on the surrounding entities. The architecture is based on a pre-trained and fine-tuned language model, where the input context is augmented with entities marked at different levels, their positions, their types, and, finally, the argument roles. Experiments on the ACE~2005 corpus demonstrate that the proposed paradigm is a viable solution for the ED task and it significantly outperforms the state-of-the-art models. Moreover, we  prove that our methods are also able to extract unseen event types. 

\end{abstract}

\section{Introduction}

Event detection (ED) is a crucial and challenging sub-task of event extraction that aims at identifying the instances of specified types of events in a text. For instance, according to the ACE 2005 annotation guidelines\footnote{\url{https://www.ldc.upenn.edu/sites/www.ldc.upenn.edu/files/english-events-guidelines-v5.4.3.pdf}}, in the sentence ``\textit{Anti-war protests took place around the world Saturday after the start of the \underline{\textbf{bombing}} in Baghdad.}'', an event detection system should be able to recognize the word \textit{bombing} as a trigger for the event \textit{Attack}.

While there are recent advances in casting the event extraction task as a machine reading comprehension (MRC\footnote{In one view, the recent tasks titled MRC can also be seen as the extended tasks of question answering (QA).}) task \cite{du2020event,liu2020event}, these models mostly focus on argument extraction, while for event detection, the models remain formulated as a sequential classification problem that aims at detecting event triggers of specific types. 

Thus, in this paper, we first propose to cast ED as a QA task with the possibility of multiple answers, in the case where more than one event is present in the text. By approaching it as a QA model, not only are we able to leverage the recent advances in MRC,  we also avoid the classification based-methods that can either require lots of training data and are challenged by the annotation cost or data scarcity. 

Second, we aim at approaching the trigger ambiguity in event detection. The same event trigger might represent different events in different contexts. For example, \textit{transfer} could refer to transferring ownership of an item, transferring money, or transferring personnel from one location to another. Each sense of the word is linked with an event type. In the same manner, \textit{fired} can correspond to an \textit{Attack} type of event as in ``\textit{The Marines \underline{\textbf{fired}} back at the gunmen} or it can express the dismissal of an employee from a job as in \textit{Offices in New Windsor and immediately \underline{\textbf{fired}} 17 staff members}''. To approach this issue, we take advantage of entities present in a sentence, considering that informative features can be brought by additional entity markers for better distinguishing the event triggers. In addition, modeling the task as QA can improve the event detection task in regards to this challenge due to the fact that the answers are only considered in relation to the context and the question, which could reduce trigger ambiguity.

Our proposed method with entity information obtains state-of-the-art results, while without it, it only obtains promising results. These methods could foster further research and help to study transfer learning from QA models to boost the performance of existing information extraction systems. Furthermore, compared to classification based-methods that lack this ability, we demonstrate that our proposed QA models are more effective in few-shot scenarios by showing that they are able to extract unseen event types.

In this paper, we prove that the ED task can be framed as a QA task and that this model can be improved by adding extra information by explicitly marking the entities in a text (entity positions, types, and argument roles). We continue with the presentation of the related work in Section \ref{section:related_work}, and we present in detail the QA model and the event trigger question templates in Section \ref{section:QA_model}. Next, we describe the entity marking in the input for the QA model in Section \ref{section:entity_model}. The experimental setup and the results are presented in Section \ref{section:experiments}. We provide a discussion of the results by analyzing the output in Section \ref{section:discussion}, followed by the conclusions and some perspectives in Section \ref{section:conclusions}. 




\section{Related Work}\label{section:related_work}

\paragraph{Event Detection}
Recently, several approaches for the event detection task that include contextual sub-word representations have been proposed, based generally on bidirectional encoder representations from Transformers (BERT). The approach attempted by \newcite{yang2019exploring} is based on the BERT model with an automatic generation of labeled data by editing prototypes and filtering out the labeled samples through argument replacement by ranking their quality. Although this paper is also based on BERT, the comparison is not straightforward as the correctness of each predicted event mention did not follow standard setups and we could not replicate their results\footnote{To the best of our knowledge, there is no public implementation and our attempt to implement their model did not achieve comparable results.}. A similar framework was proposed by \newcite{wang2019adversarial} where the informative features are encoded by BERT and a convolutional neural network (CNN), which would suggest a growing interest not only in language model-based approaches but also in adversarial models. The model proposed by \newcite{wadden2019entity} is a BERT-based architecture that models text spans and is able to capture within-sentence and cross-sentence context. 

\paragraph{Event Detection as Question Answering}
A recent work proposed by \newcite{du2020event} introduced this new paradigm for event extraction by formulating it as a QA task, which extracts the event triggers and arguments in an end-to-end manner. For detecting the event, they considered an approach based on BERT that is usually applied to sequential data. The task of ED is a classification-based method where the authors designed simple fixed templates as in \textit{what is the trigger}, \textit{trigger}, \textit{action}, \textit{verb}, without specifying the event type. For example, if they chose \textit{verb} template, the input sequence would be: {[CLS] \textit{verb} [SEP] sentence [SEP]}. Next, they use a sequential fine-tuned BERT for detecting event trigger candidates.

Another recent paper \cite{liu2020event} also approaches the event extraction task as a question answering task, similar to the \cite{du2020event} method. The task remains classification-based (instead of the span-based QA method) for trigger extraction, jointly encode {[EVENT]} with the sentence to compute an encoded representation, as in the approach proposed by \newcite{du2020event} where the special token was \textit{verb} or \textit{trigger}.

\paragraph{Question Answering}
While QA for event detection is roughly under-researched, Transformer-based models have led to striking gains in performance on MRC tasks recently, as measured on the SQuAD v1.1\footnote{SQuAD v1.1 consists of reference passages from Wikipedia with answers and questions constructed by annotators after viewing the passage} \cite{rajpurkar2016squad} and SQuAD v2.0\footnote{SQuADv2.0 augmented the SQuAD v1.1 collection with additional questions that did not have answers in the referenced passage.} \cite{rajpurkar2018know} leaderboards.

\paragraph{Entity Markers}

In the context of event detection, some works made use of gold-standard entities in different manners. Higher results can be obtained with gold-standard entity types \cite{nguyen2015event}, by concatenating randomly initialized embeddings for the entity types. A graph neural network (GNN) based on dependency trees \cite{nguyen-aaai-18} has also been proposed to perform event detection with a pooling method that relies on entity mentions aggregation. Arguments provided significant clues to this task in the supervised attention mechanism proposed to exploit argument information explicitly for ED proposed by \newcite{liu-acl-17}. Other methods that took advantage of argument information were joint-based approaches. The architecture adopted by \newcite{liu2018jointly} was to jointly extract multiple event triggers and event arguments by introducing syntactic shortcut arcs derived from the dependency parsing trees. 

In this paper, we were inspired by a recent work applied in relation extraction. \newcite{soares2019matching} experimented with entity markers for the relation extraction task by studying the ability of the Transformer-based neural networks to encode relations between entity pairs. They identified a method of representation based on marking entities that outperform previous work in supervised relation extraction. 

\section{Event Question Answering Model}\label{section:QA_model}


We formulate the ED task as a QA task, where, for every sentence, we ask if an event type of interest is present and we expect a response with an event trigger, multiple event triggers, or none. Our model extends the BERT \cite{devlin2018bert} pre-trained model which is a stack of Transformer layers \cite{vaswani2017attention} that takes as input a sequence of subtokens, obtained by the WordPiece tokenization  \cite{wu2016google} and produces a sequence of context-based embeddings of these subtokens.
 
\begin{figure}[ht]
	\centering
    \includegraphics[width=.5\textwidth]{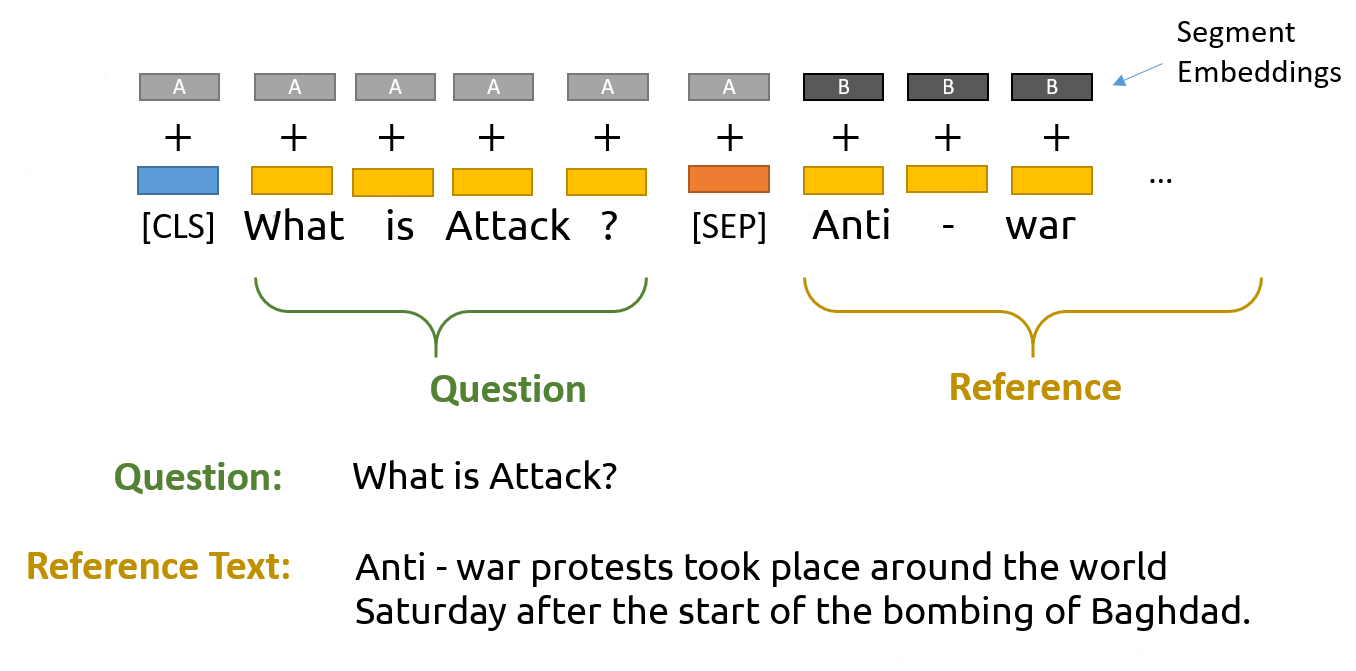}
    \caption{Example of input modification to fit the QA paradigm for a sentence that contains an event of type Attack.}
    \label{figure:bert_qa}
\end{figure}

To feed a QA task into BERT, we pack both the question and the reference text into the input, as illustrated in Figure \ref{figure:bert_qa}. The input embeddings are the sum of the token embeddings and the segment embeddings. The input is processed in the following manner: token embeddings (a [CLS] token is added to the input word tokens at the beginning of the question and a [SEP] token is inserted at the end of both the question and the reference text) and segment embeddings (a marker indicating the question or the reference text is added to each token). This allows the model to distinguish between the question and the text. 


To fine-tune BERT for a QA system, a start vector and an end vector are introduced. A linear layer is added at the top of BERT layers with two outputs for the start and end vectors of the answer. The probability of each word being the start or end word is calculated by taking a dot product between the final embedding of the word and the start or end vector, followed by a softmax over all the words. The word with the highest probability value is considered. This method differs from the event detection approaches presented by \newcite{du2020event} and \newcite{liu2020event} where the models are classification-based, instead of the span-based QA.

Next, for every type of event (\textit{Demonstrate, Die, Attack}, etc.), we formulate the question by automatically generating them using the following template:

\begin{center}
\textbf{What is the [Event Type] ?} 
\end{center}

An example for a sentence containing an \textit{Attack} event is illustrated in Figure \ref{figure:bert_qa}. We also consider questions that do not have an answer in the case where an event of a specific type is not present in the sentence. When there is more than one event of the same type in a sentence, we consider that the question has multiple answers. From the $n$ best-predicted answers, we consider all those that obtained a probability higher than a selected threshold. 

The strategy of the threshold selection is represented by the Algorithm \ref{alg:thresh_selection}, algorithm slightly similar to the method proposed by \newcite{du2020event} for determining the number of arguments to be extracted for each role by finding a dynamic threshold. When the predicted chunks are self-contained as for example, the noun chunks \textit{assault} and \textit{air assault} are predicted, we consider only the first predicted event trigger (\textit{assault}). 

\begin{algorithm}
    
    \textbf{Input:}\\ Development candidates $dev\_candidates$ \\ Test candidates $test\_candidates$ \\
    
    $list\_thresh\gets \{0.1, 0.2, 0.3, 0.4, 0.5\}$ \\
    $best\_thresh\gets 0.0$ \\
    $best\_F1\gets 0.0$\\
    \For{$thresh \in list\_thresh$}{
        $F1\gets eval(dev\_candidates)$ \\
        \If {$F1\geq$ best\_F1}{
            $best\_F1\gets F1$\\
            $best\_thresh\gets thresh$ 
        }
    }
    
    $final\_triggers\gets \{\}$ \\
    \For{$candidate \in test\_candidates$}{
        \If {$candidate.probability\geq best\_thresh$}{
            $final\_triggers.add(candidate)$\\
        }
    }
    \textbf{Output:} $final\_triggers$ \\
    \caption{Threshold selection for obtaining the top event triggers.}\label{alg:thresh_selection}
    
\end{algorithm}

\section{Event Entity and Argument Markers}\label{section:entity_model}

Next, for adding entity information, we implemented the BERT-based model with \textit{EntityMarkers[CLS]} version of \newcite{soares2019matching} applied for relation classification. 
The \textit{EntityMarkers[CLS]} model \cite{soares2019matching} consists in augmenting the input data with a series of special tokens. Thus, if we consider a sentence $x=[x_0, x_1, \ldots, x_n]$ with $n$ tokens, we augment $x$ with two reserved word pieces to mark the beginning and the end of each event argument mention in the sentence.

We propose three types of markers: (1) \textit{Entity Position Markers}, e.g. $<$E$>$ and $<$/E$>$ where $E$ represents an entity of any type, (2) \textit{Entity Type Markers}, e.g. $<$PER$>$ and $<$/PER$>$ where PER represents an entity of type Person, and (3) if the event argument roles are known beforehand, \textit{Argument Role Markers}, e.g. $<$Agent$>$, $<$/Agent$>$ where Defendant is an event argument role. We modify $x$ in the sentence \textit{\textbf{Police} have arrested \textbf{four people} in connection with the killings.}, where \textit{Police} has the argument role of an {Agent} and \textit{four people} is an argument of type {Person}: 

(1) \textit{$<$E$>$ \textbf{Police} $<$/E$>$ have arrested $<$E$>$ \textbf{four people} $<$/E$>$ in connection with the killings.} 

(2) \textit{$<$PER$>$ \textbf{Police} $<$/PER$>$ have arrested $<$/PER$>$ \textbf{four people} $<$/PER$>$ in connection with the killings.}

(3) \textit{$<$Agent$>$ \textbf{Police} $<$/Agent$>$ have arrested $<$Person$>$ \textbf{four people} $<$/Person$>$ in connection with the killings.}



Further, an event detection system should detect in the presented sentence, the trigger word \textbf{\underline{killings}} for an event of type Die (this event has two arguments \textit{Police} and \textit{four people}) and \textbf{\underline{arrested}} for an event of type Arrest-Jail (this event has no arguments). For the \textit{Argument Role Markers}, if an entity has different roles in different events that are present in the same sentence, we mark the entity with all the argument roles that it has.

\section{Experiments}\label{section:experiments}

\begin{table*}[ht]
\centering
    \begin{tabular}{|p{11.5cm}lll|} \hline
\textbf{Approaches}  & \textbf{P} & \textbf{R}  & \textbf{F1} \\ \hline
BERT-base + LSTM \cite{wadden2019entity} & N/A & N/A & 68.9 \\
BERT-base  \cite{wadden2019entity} & N/A & N/A & 69.7 \\
BERT-base \cite{du2020event} & 67.1 & 73.2 & 70.0 \\
BERT\_QA\_Trigger \cite{du2020event} & 71.1 & 73.7 & 72.3 \\
DMBERT \cite{wang2019adversarial} & 77.6 & 71.8 & 74.6 \\
RCEE\_ER \cite{liu2020event}$^+$ & 75.6 & 74.2 & 74.9 \\
DMBERT+Boot \cite{wang2019adversarial} & 77.9 & 72.5 & 75.1 \\
\hline
BERT-QA-base-cased + \textit{Entity Position Markers}$^+$ & 74.9 & 72.4 & 73.6* \\ 
BERT-QA-base-cased + \textit{Entity Type Markers}$^+$ & 76.3 & 72.2 & 74.2 \\
BERT-QA-base-cased + \textit{Argument Role Markers}$^+$ & 79.8 & 75.0 & 77.3* \\
\hline
BERT-QA-base-uncased + \textit{Entity Position Markers}$^+$ & 78.0 & 70.7 & 74.2* \\ 
BERT-QA-base-uncased + \textit{Entity Type Markers}$^+$ &  78.5 &  77.2 & 77.8* \\
BERT-QA-base-uncased + \textit{Argument Role Markers}$^+$ & \bf 83.2 & \bf 80.5 & \bf 81.8* \\ 
\hline
    \end{tabular}
    
\caption{Evaluation of our models and comparison with state-of-the-art systems
for event detection on the blind test data. $^+$ with gold entities or arguments. Statistical significance is measured with McNemar's test. * denotes a significant improvement over the previous model at p $\leq$ 0.01.
\label{table:all_results}}
\end{table*}

\begin{table}[ht]
\centering
    \begin{tabular}{|llll|} \hline
\textbf{Pre-trained BERT}  & \textbf{P} & \textbf{R}  & \textbf{F1} \\ \hline

base-cased-squad2 & 69.6 & 68.1 & 68.9 \\
base-uncased-squad2 & \bf 70.6 & 66.7 & 68.6  \\ 
\hline
base-cased & 62.2 & \bf 74.3 & 67.7 \\ 
base-uncased &  68.4 & 70.5 & \bf 69.4 \\ \hline
\end{tabular}
    
\caption{Evaluation of several pre-trained BERT-based models.\label{table:bert_models}}
\end{table}

The evaluation is conducted on the ACE 2005 corpus provided by ACE program\footnote{\url{https://catalog.ldc.upenn.edu/LDC2006T06}}. 
For comparison purposes, we use the same test set with 40 news articles (672  sentences), the same development set with 30 other documents (863 sentences) and the same training set with the remaining 529 documents (14,849 sentences) as in previous studies of this dataset \cite{ji2008refining,liao2010using}. The ACE 2005 corpus has $8$ types of events, with $33$ subtypes (e.g. the event type \textit{Conflict} has two subtypes \textit{Attack, Demonstrate}). In this paper, we refer only to the subtypes of the events, without diminishing the meaning of main event types.

\paragraph{Evaluation Metrics} Following the same line of previous works, we consider that 
a trigger is correct if its event type, subtype, and offsets match those of a reference trigger. We use Precision (P), Recall (R), and F-measure (F1) to evaluate the overall performance.  

\paragraph{Hyperparameters} We used the Stanford CoreNLP toolkit\footnote{\url{http://stanfordnlp.github.io/CoreNLP/}} to pre-process the data, including tokenization and sentence splitting. For fine-tuning the BERT-based models, we followed the selection of hyperparameters presented by \newcite{devlin2018bert}. We found that $\mathrm{3\times10}^{-5}$ learning rate and a mini-batch of dimension $12$ for the \textit{base} models provided stable and consistent convergence across all experiments as evaluated on the development set. The maximum sequence length is set to $384$ and the document stride of $128$. For selecting the event triggers, we generate $n=10$ candidates and we use the same threshold for all the experiments with a value of $0.2$ that was decided on the development set using Algorithm \ref{alg:thresh_selection}.


\paragraph{General Evaluation} We first experiment with several BERT-based pre-trained models. First, we consider the models trained on SQuAD 2.0 tailored for the extractive QA downstream task (base-cased-squad2\footnote{\url{https://huggingface.co/deepset/bert-base-cased-squad2}} and bert-uncased-squad2\footnote{\url{https://huggingface.co/twmkn9/bert-base-uncased-squad2}}), and the more general BERT models trained on large amounts of data and frequently used in research. The results are reported in Table \ref{table:bert_models}, where we can easily observe that the \textit{BERT-base-uncased} obtains the highest values. The \textit{large} BERT-based models were not considered due to memory constraints\footnote{Reducing the size of some hyperparameters for the \textit{large} models, as well as reducing the size of the batch size, decreased considerably the performance. The F1 value even plateaued at 0\%. We ran the models on a machine with four GeForce RTX 2080 GPUs, with 11,019 MiB each.}. We also distinguish between the \textit{cased} and \textit{uncased} models, where the \textit{squad2} F1 values are marginally close, as well as for the \textit{base} models.
  
\begin{figure*}[ht]
        \centering
        \includegraphics[width=.45\linewidth]{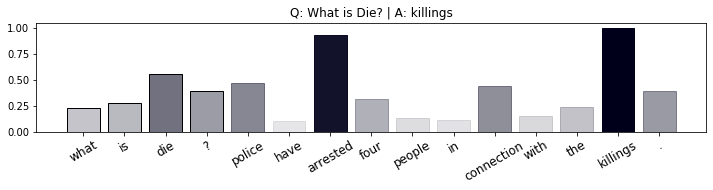}
        \includegraphics[width=.45\linewidth]{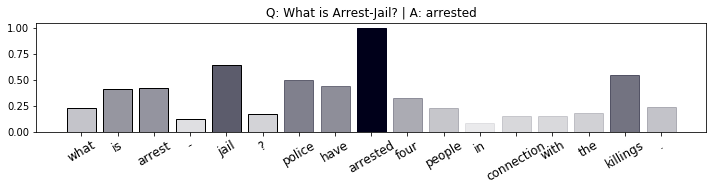}
        \caption{An example of a sentence that contains two events: \textit{Die} event triggered by the word \textbf{\underline{killings}} and \textit{Arrest-Jail} event triggered by \textbf{\underline{arrested}}. The model used is BERT-QA-base-uncased.}\label{fig:example_two_events}
\end{figure*}

In Table \ref{table:all_results}, we present the comparison between our model and the latest state-of-the-art approaches. We only compare our methods with recently proposed BERT-based models. First, we compare with the models where the task has been approached as a QA task but still formulated as a sequential classification problem that aims at locating trigger candidates, the fine-tuned baseline BERT-base-uncased  \cite{du2020event}, the BERT\_QA\_Trigger \cite{du2020event}, and the RCEE\_ER (\underline{R}eading \underline{C}omprehension for \underline{E}vent \underline{E}xtraction, with \textit{ER} that denotes that the model has golden entity refinement) \cite{liu2020event}.

We also compare with the two models with adversarial training for weakly supervised event detection proposed by \newcite{wang2019adversarial}, the BERT and LSTMs approaches proposed by \newcite{wadden2019entity} that models text spans and captures within-sentence and cross-sentence context, and DMBERT \cite{wang2019adversarial} with adversarial training for weakly supervised ED. 


When compared with the BERT\_QA\_Trigger \cite{du2020event}, our models that use either the positions or the types of the entities bring a considerable improvement in the performance of trigger detection. It is clear that further marking the entities with their types can increase both precision and recall, balancing the final scores.

From Table \ref{table:all_results}, one can observe that, on the diagonal, the most balanced models with regard to the precision and the recall values  mainly consist in the models that include either \textit{Entity Type} or \textit{Argument Role Markers}, along with the QA classification-based models proposed by \newcite{du2020event,liu2020event}. 

For the BERT-QA-base-uncased + \textit{Entity Position Markers} model and the DMBERT-based models \cite{wang2019adversarial}, the results are visibly imbalanced, with high precision values, which implies that these models are more confident in the triggers that were retrieved. Moreover, marking the entities with $<$E$>$ and $<$/E$>$ and BERT-QA-base-uncased has the lowest values in recall, but adding \textit{Position Markers} clearly increases the precision of the results.

While entities can be present in the entire document, arguments can only surround event triggers. Knowing the argument roles beforehand brings further improvements, and we assume that an important reason for this is that, since the arguments are present only around event triggers, this could help the language model to be more aware of the existence of an event or multiple events in a sentence.

\begin{figure}[ht]
        \centering
        \includegraphics[width=1.\linewidth]{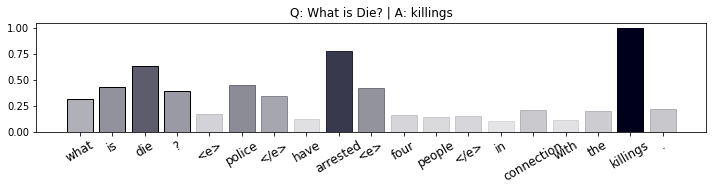}
        \includegraphics[width=1.\linewidth]{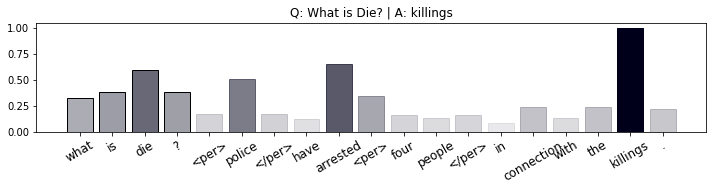}
        \includegraphics[width=1.\linewidth]{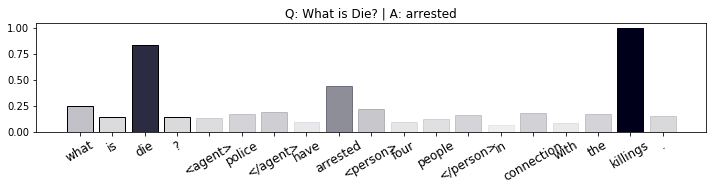}
        \caption{An example for the \textit{Die} event triggered by \textbf{\underline{killings}} with three types of markers: \textit{Entity Position, Entity Type}, and \textit{Argument Role Markers}.}\label{fig:example_all_models}
\end{figure}
\section{Discussion}\label{section:discussion}

For a deeper analysis of the impact of entity information, we leverage the gradients in our proposed models to efficiently infer the relationship between the question, context, and the output response. \newcite{brunner2019identifiability} studied the identifiability of attention weights and token embeddings in Transformer-based models. They show that the self-attention distributions are not directly interpretable and suggest that simple gradient explanations are stable and faithful to the model and data generating process. 
Thus, as applied by \newcite{madsen2019visualizing}, to get a better idea of how well each model memorizes and uses memory for contextual understanding, we analyze the connectivity between the desired output and the input. This is calculated as:  $$
\mathrm{connectivity}(t, \tilde{t}) = \left|\left| \frac{\partial y^{\tilde{t}}_{k}}{\partial x^t} \right|\right|_2
$$  where $t$ is the time index, $\tilde{t}$ the output time index, and the result is the magnitude of the gradient between the logits for the desired output $y^{\tilde{t}}_{k}$ and the input $x^t$. The connectivity is computed with respect to both start position and end position of the answer, then it is normalized, and it is visible as saliency maps for every word in Figures \ref{fig:example_two_events} and \ref{fig:example_all_models}\footnote{The sentence is lowercased for the \textit{uncased} models.}.

By looking at the gradients in Figure \ref{fig:example_two_events}, where two events of different types are present, we can observe, in the upper part of the figure, that while the model sees the word \textbf{\underline{killings}} and \textbf{\underline{arrested}} as impactful, it also sees the words \textit{police, connection} as impactful and selects an answer in that neighborhood. Even though both trigger candidates \textbf{\underline{killings}} and \textbf{\underline{arrested}} have a clear impact due to their gradient values, by looking at the probability values, \textbf{\underline{killings}} is recognized with a $99.4$\% probability, while \textbf{\underline{arrested}} obtained a probability of $2.3\times10^{-7}$, value that is lower than our selected threshold $0.2$. In the lower part of the figure, for the question \textit{What is Arrest-Jail?}, the words \textit{die, police,} \textbf{\underline{killings}} clearly influence the choice of the answer \textbf{\underline{arrested}}.


In Figure \ref{fig:example_all_models}, we present the same sentence with the three types of input modifications: \textit{Entity Position Markers, Entity Type Markers}, and \textit{Argument Role Markers}, with the \textit{What is Die?} question and the correct answer \textbf{\underline{killings}}. In the upper part of the figure, where the sentence has been augmented with the entity position markers $<$E$>$ and $<$/E$>$, we notice that the words that impact the most the result are \textbf{\underline{killings}} along with \textit{die}, \textbf{\underline{arrested}}, and \textit{police}. In this case, one can also see that the end marker $<$/E$>$ contributed too. 

\begin{figure*}[ht]
        \centering
        \includegraphics[width=0.22\linewidth]{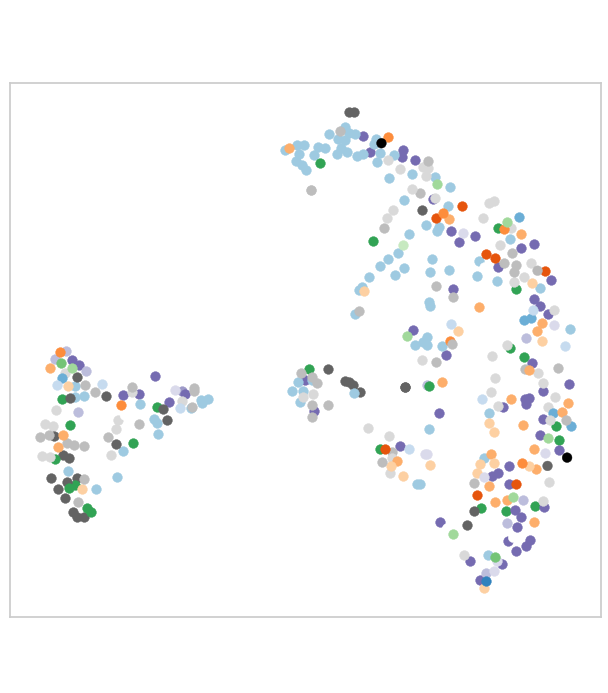}
        \includegraphics[width=0.2044\linewidth]{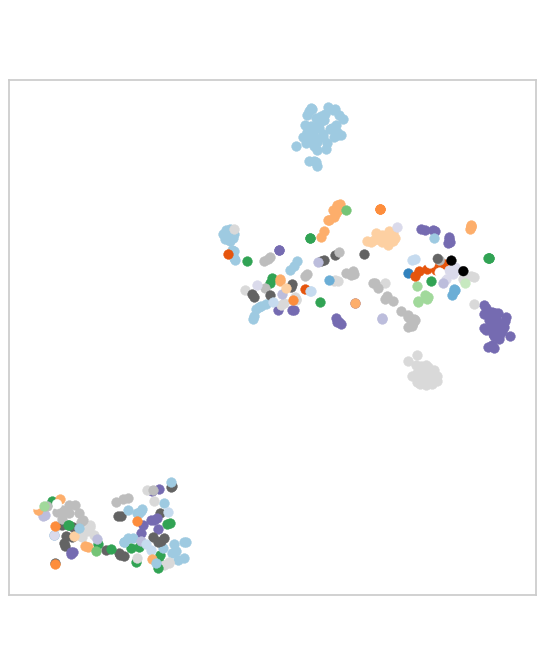}
        \includegraphics[width=0.22\linewidth]{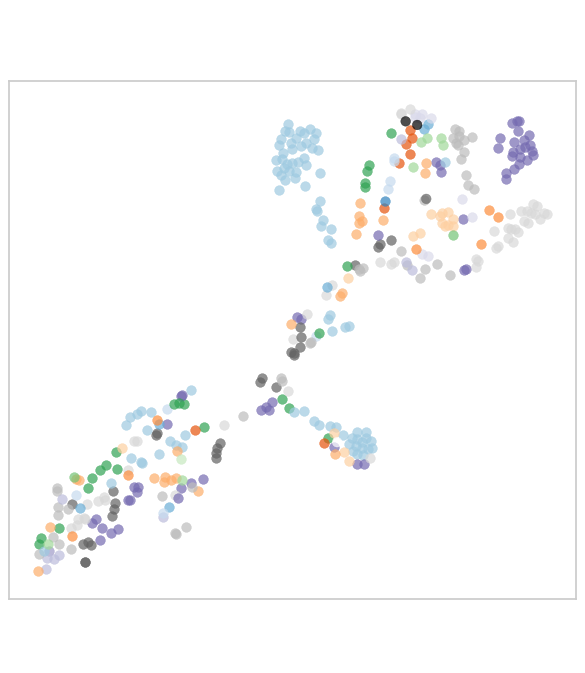}
        \includegraphics[width=0.296\linewidth]{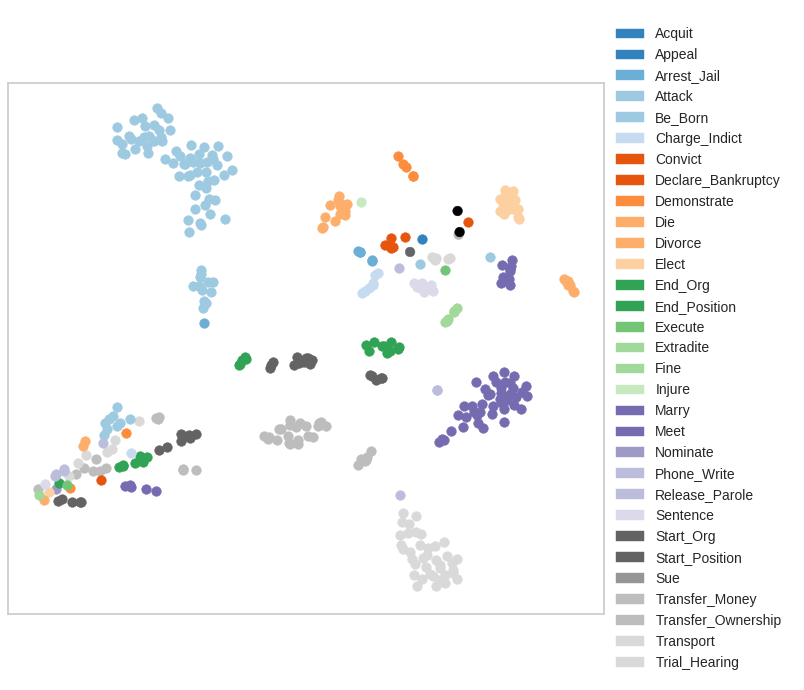}
        \caption{{[CLS]} representation of each sentence in the test set that contains at least an event for BERT-QA-base-uncased, BERT-QA-base-uncased +\textit{ Entity Position Markers+}, BERT-QA-base-uncased + \textit{Entity Type Markers+}, and  BERT-QA-base-uncased + \textit{Argument Role Markers}.}\label{fig:sentence_representations}
\end{figure*}

In the middle part of the figure, where the sentence has been augmented with the entity position markers $<$PER$>$ and $<$/PER$>$ for the two entities \textit{police} and \textit{four people}, the influence of other words as in \textit{die}, \textbf{\underline{arrested}}, and \textit{police} slightly decreased. In the bottom part of the image, the gradients of these words are visibly reduced. 

When the sentence is augmented with argument roles, $<$Agent$>$, $<$/Agent$>$, $<$Person$>$ and $<$/Person$>$, the noise around the correct answer has noticeably diminished, being reduced by the additional markers. The most impactful remaining words are the word \textit{die} in the question and the correct answer \textbf{\underline{killings}}.

In order to analyze the quality of the sentence representations, we extract the {[CLS]} representation of each sentence for BERT-QA-base-uncased and for BERT-QA-base-uncased + \textit{Argument Role Markers}. Then, we plot these representations in two spaces where the labels (colors of the dots) are the event types, as illustrated in Figure \ref{fig:sentence_representations}. On the right-hand side of the figure, where argument role markers are used, it is clear that the sentence representations clusters are  more cohesive than when no entity information is considered (left-hand side), thus confirming our assumption regarding the importance of the entity informative features in a QA system.


\paragraph{Evaluation on Unseen Event Types} We follow the same strategy as \newcite{du2020event} where we keep 80\% of event types (27) in the training set and 20\% (6) unseen event types in the test set. More exactly, the unseen event types were chosen randomly and they are: \textit{Marry, Trial-Hearing, Arrest-Jail, Acquit, Attack}, and \textit{Declare-Bankruptcy}. Table \ref{table:unseen_event_types} presents the performance scores of our models for the unseen event types. 

We compare with BERT-QA-Baseline which is our baseline that selects an event trigger in a sentence without being trained on ACE~2005 data. Since the models proposed for ED by \newcite{du2020event} and \newcite{liu2020event} are classification-based in a sequential manner, they are not capable of handling unseen event types. 

From the results, without any event annotation, the BERT-QA-base-uncased-Baseline obtains a low F1 value (1.3\%). We observe that the performance values increase proportionally to the specificity of the markers. Thus, it is not surprising that the highest values are obtained when the argument roles are marked, also obtaining the highest precision. These results also confirm the effectiveness of our proposed models in handling unseen event types. 

\begin{table}[ht]
\centering
    \begin{tabular}{|p{4.2cm}p{.6cm}p{.6cm}p{.7cm}|} \hline
  \textbf{Approaches}  & \textbf{P} & \textbf{R}  & \textbf{F1} \\ \hline
  \begin{tabular}{p{4.2cm}}
   BERT-QA-base-uncased-Baseline \\
  \end{tabular} & 0.7 & 8.3 & 1.3 \\ \hline
  \begin{tabular}{p{4.2cm}}
   BERT-QA-base-uncased\\
  \end{tabular} & 47.7 & 26.7 & 31.1 \\  \hline
  \begin{tabular}{p{4.2cm}}
    BERT-QA-base-uncased + \textit{Entity Position Markers}$^+$ \\
  \end{tabular} & 44.0 & 47.5 & 37.3 \\  \hline
  \begin{tabular}{p{4.2cm}}
    BERT-QA-base-uncased + \textit{Entity Type Markers}$^+$ \\
  \end{tabular} & 53.6 & \bf 54.4  & 50.4  \\  \hline
  \begin{tabular}{p{4.2cm}}
    BERT-QA-base-uncased + \textit{Argument Role Markers}$^+$ \\
  \end{tabular}
& \bf 83.3 & 47.4 & \bf 53.6 \\ 
\hline
    \end{tabular}
\caption{Evaluation of our models on unseen event types. $^+$ with gold entities or arguments. \label{table:unseen_event_types}}
\end{table}

\section{Conclusions and Perspectives}\label{section:conclusions}

In this paper, we proposed a new technique for detecting events by modeling the problem as a question-answering task. The questions are simplified to a pre-defined list with a question for every type of event present in the dataset, which allows the model to  predict multiple types of events in a sentence including unseen event types. 

The additional informative features brought by the presence of entities and the argument roles in the same context of the events considerably increased the performance of the model, achieving state-of-the-art results. 

In future work, we will focus on approaching the entity and argument detection task, in order to analyze the influence of the predicted event participants and the error propagation from this task to the downstream event detection task.

 \bibliographystyle{acl_natbib}
 \bibliography{acl2021}

\end{document}